\documentclass[sigconf]{acmart}
\renewcommand\footnotetextcopyrightpermission[1]{} 
\usepackage{multirow}
\usepackage{balance}
\usepackage{algorithm}
\usepackage{algpseudocode}
\usepackage{amsmath}
\usepackage{algorithmicx}
\usepackage{algpseudocode}
\usepackage{float}
\usepackage{hyperref}

\AtBeginDocument{%
  \providecommand\BibTeX{{%
    \normalfont B\kern-0.5em{\scshape i\kern-0.25em b}\kern-0.8em\TeX}}}

\copyrightyear{2021} 
\acmYear{2021} 
\setcopyright{acmcopyright}\acmConference[MM '21]{Proceedings of the 29th ACM
International Conference on Multimedia}{October 20--24, 2021}{Virtual Event, China}
\acmBooktitle{Proceedings of the 29th ACM International Conference on Multimedia
(MM '21), October 20--24, 2021, Virtual Event, China}
\acmPrice{15.00}
\acmDOI{10.1145/3474085.3475640}
\acmISBN{978-1-4503-8651-7/21/10}

\begin{document}

\title{Pixel-wise Graph Attention Networks for Person Re-identification}


\author{Wenyu Zhang,\quad Qing Ding*,\quad Jian Hu,\quad Yi Ma,\quad Mingzhe Lu}

\makeatletter
\def\authornotetext#1{
\if@ACM@anonymous\else
    \g@addto@macro\@authornotes{
    \stepcounter{footnote}\footnotetext{#1}}
\fi}
\makeatother

\authornotetext{Corresponding author.}

\affiliation{
 \institution{University of Science and Technology of China}
  \city{Hefei}
  \state{AnHui}
  \country{China}
 }
\email{{wenyuz, hujian1, my1996mh, mingzhe}@mail.ustc.edu.cn, dingqing@ustc.edu.cn}

\def\authors{Wenyu Zhang, Qing Ding, Jian Hu, Yi Ma, Mingzhe Lu}

\begin{abstract}
  Graph convolutional networks (GCN) is widely used to handle irregular data since it updates node features by using the structure information of graph. With the help of iterated GCN, high-order information can be obtained to further enhance the representation of nodes. However, how to apply GCN to structured data (such as pictures) has not been deeply studied. In this paper, we explore the application of graph attention networks (GAT) in image feature extraction. First of all, we propose a novel graph generation algorithm to convert images into graphs through matrix transformation. It is one magnitude faster than the algorithm based on K Nearest Neighbors (KNN). Then, GAT is used on the generated graph to update the node features. Thus, a more robust representation is obtained. These two steps are combined into a module called pixel-wise graph attention module (PGA). Since the graph obtained by our graph generation algorithm can still be transformed into a picture after processing, PGA can be well combined with CNN. Based on these two modules, we consulted the ResNet and design a pixel-wise graph attention network (PGANet). The PGANet is applied to the task of person re-identification in the datasets Market1501, DukeMTMC-reID and Occluded-DukeMTMC (outperforms state-of-the-art by 0.8\%, 1.1\% and 11\% respectively, in mAP scores). Experiment results show that it achieves the state-of-the-art performance. \href{https://github.com/wenyu1009/PGANet}{The code is available here}.
\end{abstract}




\maketitle

\section{Introduction}
Person re-identification attracts the attention of scholars. These studies mainly use the image of a specific pedestrian as the query to find all the images of the pedestrian in the datasets (gallery). The main application fields of it include video surveillance, intelligent security and so on. There are many challenges in person re-identification tasks: limited training data, occlusion, lighting, scale, pose, and so on. How to obtain robust features of the image is the key to solving these challenges.

Convolutional neural networks (CNN) has been successfully applied in the field of person re-identification. Existing methods mainly adopt ResNet \cite{2016Deep} as the backbone to extract the features of images. However, due to the limitation of CNN’s receptive field, the CNN based methods are difficult to identify blurred pictures or small targets. To solve above problems, it is necessary to give higher weight to human body information in the feature extraction stage and extend the receptive field to obtain global information. 

Since the attention mechanism \cite{2017Attention} can get the correlation between objects, and generate different weights. CNN with attention mechanism \cite{2017Non,2017Squeeze,2018CBAM} is a good way to solve the above problems. In the field of person re-recognition, Some typical models \cite{2019Self,2020Relation} apply attention mechanism to the feature maps output by CNN to calculate the importance of each feature in feature maps and assign different weights to highlight important features and suppress unimportant features. However, the existing methods only get the relationship between nodes as one-order relationship \cite{2019Self} or global relationship \cite{2020Relation}, which can't obtain the implicit high-order relationship.

To expand the receptive field, dilated convolutions can be applied in the feature extraction stage \cite{2016Multi}, but the continuity of features will be lost due to the discontinuity of convolution kernel. As a result, it is difficult to obtain complete features of small objects.

Human beings get the overall cognition through the continuous expansion of a certain point to the surroundings. Inspired by this ecology, graph attention networks (GAT) \cite{2017Graph} can be used to solve the above problems by simulating the characteristics of human vision. The features of nodes are propagated on the edges given different weights, and each node obtains a new representation by aggregating the neighbor features. In this way, besides important features will be assigned a higher weight, the range of receptive field will be scalable.

\begin{figure}[htp]
  \centering
  \includegraphics[width=\linewidth]{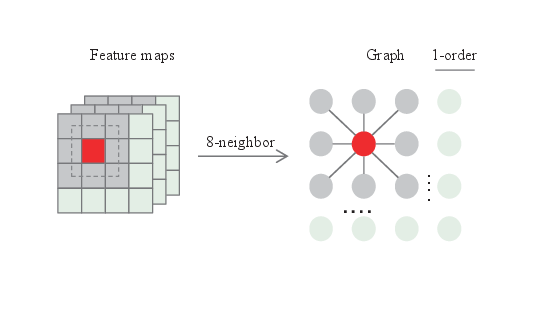}
  \caption{We treat the picture as a regular graph and optimize the features of the picture from the perspective of graph convolution.}
  \Description{picture transf to graph.}
  \label{pic}
\end{figure}

The premise of using GAT is to convert an image into a graph and obtain its adjacency matrix. As shown in Fig. \ref{pic}, a single pixel is regarded as a node, and its adjacent pixels are neighbors. Based on this setting, we propose a novel graph generation algorithm, which is not only faster than the traditional algorithm but also obtains a sparse and strictly local graph. Subsequently, using GAT on the obtained graph, not only can obtain high-order information of neighbor nodes but also get a large receptive field. These two steps are combined into a module called pixel-wise graph attention module (PGA). Combined with ResNet, we build a new model pixel-wise graph attention network (PGANet). And we apply the model to the commonly used data sets Market1501\cite{2015Scalable}, DukeMTMC-reID\cite{2016Performance} and Occluded-DukeMTMC\cite{2019Pose} for person re-identification.

In summary, the contribution of this paper is threefold:

\begin{itemize}
\item {\verb|We propose a novel graph generation algorithm|}: The algorithm can transform the image into a sparse and strictly local graph, which can make use of the structure information. It is one order of magnitude faster than the traditional KNN based algorithm.
\item {\verb|We propose a PGA module|}: Multi-layer PGA can simulate human vision. With the continuous iteration of PGA, the receptive field is continuously expanded and high-level information is obtained.
\item {\verb|We propose a new model PGANet|}: It is composed of CNN and PGA. We apply it to person re-identification to demonstrate the effectiveness of our model.
\end{itemize}

\section{Related Work}
In this section, we introduce the existing methods of image feature extraction and the application of graph convolution networks in the field of computer vision.

\subsection{Image Feature Extraction Based On CNN}
 CNN based methods for person re-identification often use ResNet \cite{2017SVDNet,2017In,2016Performance} as the backbone to extract image features. However, due to the limited receptive field, CNN can not obtain global information. As a result, it is difficult to distinguish blurred images and small targets. To solve this problem, some methods have been proposed. The part-based models \cite{2016A,2017Beyond,2017HydraPlus,2017Deeply,2018Robust,2018Horizontal,2018Learning}, divide the picture into sub-pictures to get the part-level features. So they can provide fine-grained information for better representation. But the processed images are required to be aligned. Attention-based models \cite{2019Concentrated,2019Self,2020Relation} assign different importance to different features in the feature extraction process. Therefore, it can obtain the relationship between each pixel in the picture. But, the existing methods only obtain one-order relationships or global information\cite{2020Relation} and the implicit high-order information cannot be obtained. Dilated convolutions can be applied in the feature extraction to overcome the limited receptive field problem \cite{2016Multi}, but the continuity of features will be lost due to the discontinuity of the convolution kernel. 

\subsection{Image Feature Extraction Based On GCN}

GCN aggregates neighbor node data along the direction of the edge to update the new representation of the source node. It is mainly divided into two categories: spectral domain approaches \cite{2013Spectral,cnn_graph,2016Semi} and spatial domain approaches \cite{2017Graph,2017Geometric,2017Inductive,2020DeepGCNs}. Because node neighbors can be flexibly aggregated and it has lower computational complexity, spatial GCN is the mainstream method. Hamilton et al.\cite{2017Inductive} propose the GraphSAGE framework, which efficiently generates node embeddings for previously invisible data using node feature information. In GAT model \cite{2017Graph}, attention mechanism is adopted in the propagation stage. Li et al. \cite{2020DeepGCNs} proposed that GCN can go deeper as CNN by applying residual structure to alleviate the problem of vanishing gradient. Recently, GCN has been widely used in the field of computer vision. In the task of multi-label image recognition, Chen et al. \cite{2019Multi} propose a novel reweighting scheme to create an effective label correlation matrix to guide the information dissemination between nodes in GCN. Context-aware graph convolutional network (CAGCN) \cite{2020Context}, in which the relationship of the probe-gallery is encoded into the graph, processes the hard samples through graph reasoning by using the context information flow and other simple samples. However, GCN is rarely used in image feature extraction. Defferrard et al. \cite{cnn_graph} proposed to apply a clustering algorithm to convert the image into the graph, and then use GCN in the generated graph. The computation complexity of the method to build a graph is as high as $\mathbf{O(N^2)}$. In the Dual Attention Network (DANet) model\cite{2020Dual}, attention is applied between each pixel to establish a fully connected graph, which was used to aggregate the features of nodes and applied it in image segmentation. But the graph obtained in this way connects each node to all nodes, ignoring the structural information.

\section{Methodology}

\begin{figure*}[htb]
  \centering
  \includegraphics[width=0.95\linewidth]{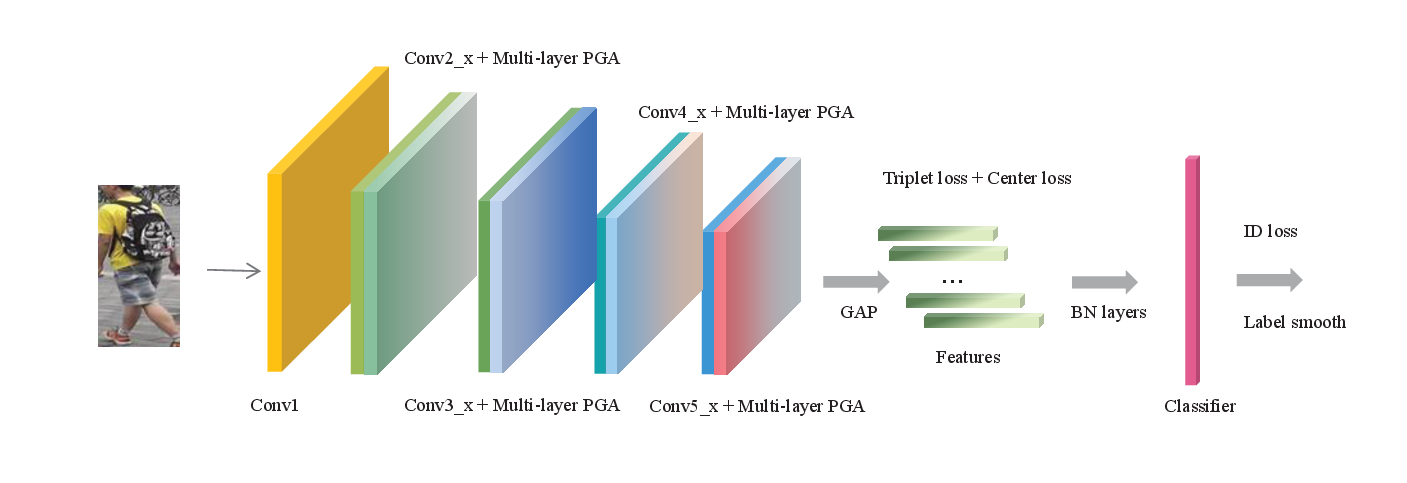}
  \caption{In PGANet, the features extracted by CNN are continuously optimized in the PGA module. Ultimately get a more robust picture representation}
  \Description{architecture}
  
  \label{arc}
\end{figure*}

PGANet is the backbone of image processing in our method. Its structure is shown in Fig. \ref{arc} and \ref{pga}. Firstly, we design a graph generation module to transform the output of CNN into a graph. Then the graph is processed by GAT. These two modules are built into a multi-layer PGA. Finally, multi-layer PGA is inserted into Conv\_x of ResNet. As shown in Fig. \ref{arc}, a new framework PGANet is constructed by stacking multi-layer PGA and CNN.

\subsection{Graph Generation Algorithm}

The graph can be formulated as $G = <V, E, A>$, in which $V$ is the set of nodes in the graph, $E$ is the edges between nodes in the graph, and the adjacency matrix $A$ is usually used to represent the connection relationship of each node. The feature maps output by CNN  is expressed as $F\in\mathbb R^{H\times W\times C}$, where H, W, and C represent the height, width, and channel respectively. We regard the feature maps as regular graph. The pixel in the feature map is regarded as the node in the graph, and its value in every channel is the feature of the node, which is a C-dimensional feature vector. 
	
\begin{equation}
A=\left\{\begin{array}{ll}1, & v_{j} \text{ is the neighbor of } v_{i} \\0, & v_{j} \text { isn't the neighbor of } v_{i}\end{array}\right.
\label{1}
\end{equation} 

As shown in Eq. \ref{1}, a graph generation algorithm is needed to generate a sparse and strictly local adjacency matrix $A$. The traditional method uses a clustering algorithm to obtain the adjacency matrix of the graph, the time complexity is as high as $\mathbf{O (N^2)}$. We propose a novel graph generation algorithm, which can efficiently find 4-neighbor, 8-neighbor, or 2-neighbor of nodes. In particular, the 2-neighbor's graph is constructed in the channel of the picture, thus each feature map is flattened as the feature of the node. Our algorithm finds neighbors of nodes in parallel and does not need to calculate the distance between nodes. The time complexity of algorithm 1 is $\mathbf{O (kHW)}$, since the $\mathbf{for}$ loop has $\mathbf{O (H)}$ complexity, the operation of $\mathbf{addition}$, $\mathbf{subtraction}$, $\mathbf{slicing}$, and $\mathbf{assignment}$ in each loop has $\mathbf{O (5kW)}$ complexity. We know that $\mathbf{N = H*W}$ and $\mathbf{k}$ is 2, 4, or 8. When $\mathbf{N}$ is very large, the time complexity of algorithm 1 is smaller than the traditional algorithm, $\mathbf{O (kHW) < < O (N ^ 2)}$. As shown in Algorithm 1, the algorithm which generates a 4-neighbor graph is described.

\begin{algorithm}[htp]
    \caption{Graph Generation Algorithm}
    \begin{algorithmic}[1] 
            \Require $\mathtt{h}$ is the height of feature maps, $\mathtt{w}$ is the width of feature maps,$\mathtt{n} = \mathtt{h}\times\mathtt{w}$ is the  number of pixels in the feature maps, \Call{AdjacencyGeneration}{$\mathbf{node}, \mathbf{neighbor}$}is a function to generate adjacency matrix based on node pairs.
        \Ensure $\mathbf{A}$ is Adjacency matrix of feature maps.
        \Function {GraphGeneration}{$\mathtt{h}, \mathtt{w},\mathtt{n}$}
            \State $\mathbf{M} \gets$  \Call{range}{$\mathtt{n}$}.\Call{reshape}{$\mathtt{h},\mathtt{w}$}
            \State $\mathbf{r}  \gets \mathbf{M}$.rows ; 
            \State $\mathbf{node} \gets [ ]$ ; $\mathbf{neighbor} \gets [ ]$;
            \For{$i = 0 \to \mathtt{h-1}$}
                \State $\mathbf{node} \gets \mathbf{r_{i}}[1:]$
                \State $\mathbf{neighbor} \gets \mathbf{(r_{i}}-1)[1:]$
                \State $\mathbf{node} \gets \mathbf{r_{i}}[:-1]$
                \State $\mathbf{neighbor} \gets \mathbf{(r_{i}}+1)[:-1]$
                \If{i != h-1} 
                    \State $\mathbf{node} \gets \mathbf{r_{i}}$
                    \State $\mathbf{neighbor} \gets \mathbf{(r_{i}}+w)$
                \EndIf
                \If{i != 0} 
                    \State $\mathbf{node} \gets \mathbf{r_{i}}$
                    \State $\mathbf{neighbor} \gets \mathbf{(r_{i}}-w)$
                \EndIf
            \EndFor
            \State $\mathbf A \gets $ \Call{AdjacencyGeneration}{$\mathbf{node}, \mathbf{neighbor}$}
            \State \Return $\mathbf A$
        \EndFunction
    \end{algorithmic}
\end{algorithm}

\subsection{Pixel-wise Graph Attention Networks}

\begin{figure*}[!ht]
  \centering
  \includegraphics[width=\linewidth]{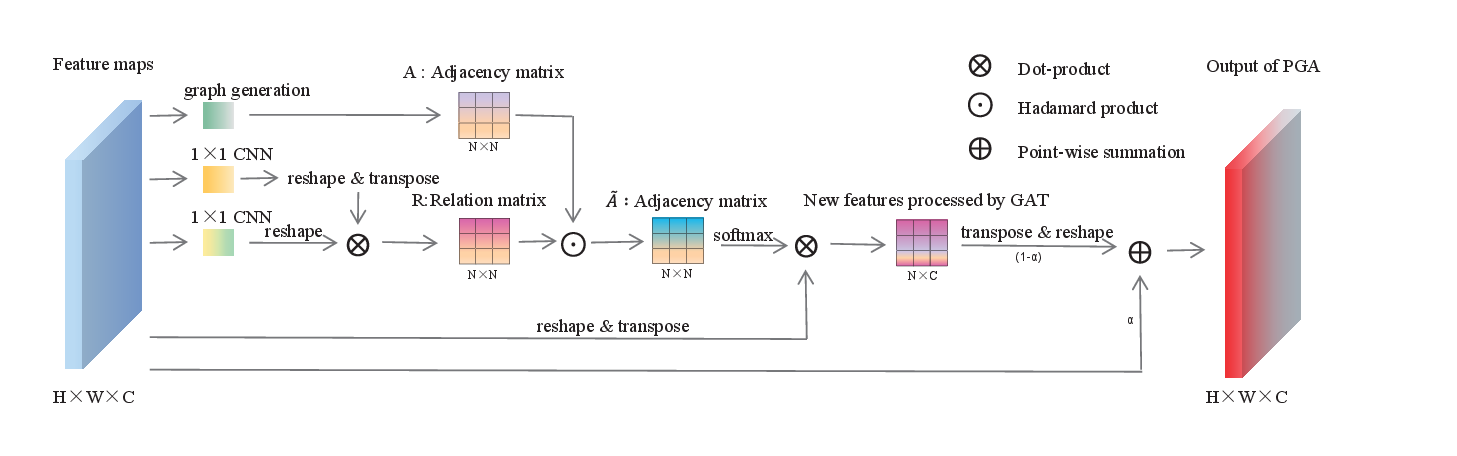}
  \caption{Diagram of our proposed pixel-wise graph attention module (PGA). PGA obtains the attention between each node through non-local operations. We use GAT to update the features of each node.}
  \Description{Detail of PGA.}
  \label{pga}
\end{figure*}

PGANet is composed of multiple PGA modules connected in series. The operation process of PGA is shown in Fig. \ref{pga}. The self-attention mechanism utilized by PGA closely follows the work of non-local operations \cite{2017Non}. In PGA, the propagation mechanism of node features is modified by GAT \cite{2017Graph}.

We first describe a single PGA. As shown in Fig. \ref{pga}, the feature maps $F\in\mathbb R^{C\times W\times H}$ output by Conv\_x in ResNet are used as the input of PGA. The corresponding adjacency matrix $A$ is generated based on the graph generation algorithm proposed in Section 3.1. Subsequently, the correlation matrix $R$ is obtained by calculating the correlation of each node in the feature maps, as shown in Eq \ref{2}.

\begin{equation}
R =\theta\left(F\right)\phi\left(F\right)^{T}
\label{2}
\end{equation}

where $(·)^{T}$ stands for transpose. $\phi (·)$ and $\theta (·)$ are the same transfer functions with different parameters, which including 1×1 convolutional layers, batch normalization, ReLU activation, and matrix transformation. The transfer function changes the feature maps to new feature space and transforms the new feature maps into a set of nodes through matrix transformation, e.g., $V^{'}=\phi(F)$, it can be defined as $V^{\prime}=\left\{v_{i}^{\prime} \in \mathbb{R}^{C^{\prime}}, i=1, \ldots, N\right\}$, where $v^{'}_i$ represents the $i^{th}$ node, $C^{'}$ is the transformed feature dimension, $N = H \times W$ is the  number of all nodes in the feature maps. As shown in the Fig. \ref{pga}, we obtain a matrix of node correlation $R \in \mathbb R^{N \times N}$ through the help of matrix change and matrix multiplication. 

The previous method \cite{2020Dual} regards this correlation matrix $R$ as a graph of feature maps. But $R$ is a fully connected graph, each node is connected to all nodes. As a result,  all structural information is dropped \cite{2017Graph}. Therefore, we map the correlation matrix $R$ to the adjacency matrix $A$. We obtain the adjacency matrix with correlation and use the softmax function to normalize it. As shown in Eq. \ref{3}:

\begin{equation}
\tilde{A}=\operatorname{softmax}(A \odot R)
\label{3}
\end{equation}

As shown in Eq.\ref{3}, different weights are assigned to all the edges on the graph. We get a sparse and strictly local adjacency matrix $\tilde A  \in \mathbb R^{N \times N}$. The next step is to propagate the node features on the graph. As shown in Eq. \ref{4}:

\begin{equation}
\tilde{V}=\operatorname{ReLU} (\tilde{A} V)
\label{4}
\end{equation}

The new feature is $\tilde{V} \in \mathbb R^{N \times C}$. By using matrix transformation, it can be converted into feature maps again. Essentially, PGA is a variant of GAT. The single-layer PGA can be expressed as:

\begin{equation}
P G A\left(F\right) = \operatorname{transf}\left(\operatorname{softmax}\left(A \odot R \right) \operatorname{transf}(F)\right).
\label{5}
\end{equation}


where the operation of function $transf (·)$ is related to the form of the input. If the form of the input data is $\mathbb R^{C\times H\times W}$, we first transpose it and then execute function $reshape (-1, C)$. If the form is $\mathbb R^{N\times C}$, we transpose it and then execute function $reshape (H,W,C)$. As same as GCN, PGA has the problems of vanishing gradient and over-smooth. We add the residual structure to the PGA with regarding to DeepGCNs \cite{2020DeepGCNs}. As shown in Eq. \ref{6}:

\begin{equation}
\tilde{F}=\alpha F+(1-\alpha) P G A\left(F\right)
\label{6}
\end{equation}

where $\alpha$ is a learnable parameter. Therefore, multi-layer PGA is shown in Eq. \ref{7}:

\begin{equation}
\tilde{F}^{l+1}=\alpha F^{l}+(1-\alpha) P G A\left(F^{l}\right)
\label{7}
\end{equation}

\subsection{Loss Function}
\begin{equation}
L=L_{I D}+L_{\text {Triplet }}+\beta L_{C}
\label{8}
\end{equation}

AS shown in Eq. \ref{8}, the loss function of the model mainly includes ID loss, triplet loss\cite{2017In}, and Center loss\cite{2016A}. Following strong baseline\cite{2019Bags}, ID Loss use label smoothing strategy.

\section{Experiment}
We introduced the implementation details of our model in Section 4.1. The datasets and evaluation metrics adopted in the experiment are introduced in section 4.2. In section 4.3, we compare our model with several state-of-the-art methods in the task of person re-identification to verify the superiority of it. In section 4.4, the effectiveness of the graph generation algorithm and multi-layer PGA module is proved through ablation study.

\subsection{Implementation Details} 
We use ResNet \cite{2016Deep} to build the model. Refer to \cite{2020Relation}, we insert multi-layer PGA module after each Conv\_x module (including Conv\_2, Conv\_3, Conv\_4 and Conv\_5). In our model, the number of layers of PGA is set to 3. The input image size is 256X128, and the padding value is set to 10. We use all the tricks summarized in Strong Baseline \cite{2019Bags}. To expand the size of the feature maps output by the model, the last spatial down-sampling operation is set to 1. The batch Standardization (BN) layer is added before the classifier. For data augmentation, we apply random cropping, horizontal flipping, and random erasing. The random probability for image horizontal flip and random erasing is set to $0.5$ respectively. In our experiments includes three losses as following: Identification loss\cite{2016Rethinking}, Triplet loss\cite{2015FaceNet} and Center loss\cite{2016A}. We adopt Adam optimizer with the weight decay of $5 \times 10^{-4}$ and the learning rate as $3e^{-4}$. The warm-up strategy is used for the learning rate, and iterations of warm-up are set to 500. Finally, at the re-rank stage, we use the k-reciprocal method\cite{2017Re}.

\subsection{Datasets and Evaluation Metrics}
We verify PGANet on the common datasets including Market1501\cite{2015Scalable}, DukeMTMC-reID\cite{2016Performance} and Occluded-DukeMTMC\cite{2019Pose}.

\textbf{Market1501} It contains 1501 pedestrians captured by 6 cameras ( 5 high-definition cameras and 1 low-definition camera) and 32668 detected pedestrian rectangular boxes. Each pedestrian is captured by at least 2 cameras and may have multiple images in one camera. The training set contains 751 people, with 12936 images, and each person has an average of 17.2 training data. In the test set, there are 750
people with 19732 images, and each person has an average of 26.3 test data.

\textbf{DukeMTMC-reID} It is a new type large high definition (HD) video data set recorded by 8 synchronous cameras. The video is sampled every 120 frames to form an image, and total 36, 411 images are obtained. A total of 1404 people appeared under more than two cameras, and 408 people appeared under only one camera. 702 people are randomly sampled into the training set and 1110 people in the test set. In the test set, one photo under each camera of each ID is sampled as a query image. The remaining images are added to the search gallery of the test, and the previous 408 people are also added to the gallery as interference items.

\textbf{Occluded-DukeMTMC} It derived from DukeMTMC-reID. In this dataset, all query images are occluded by large variety of objects (e.g., trees, cars, other persons). The training, query, and gallery set contain 14\%/15\%/10\% occluded images, respectively. The training set of Occluded DukeMTMC contains 15,618 images covering 702 identities in total. The testing set contains 1,110 identities, with 17, 661 gallery images and 2, 210 query images.

\textbf{Evaluation Metrics} Since cumulative matching characteristics (CMC) and mean average precision (mAP) are commonly used metrics for person re-identification tasks, they are also considered as evaluation indicators in this paper.

\begin{table}[htb]
\centering
\caption{Performance (\%) comparisons with the state-of-the-arts methods on Market1501 and DukeMTMC-reID.}
 \label{tab:compare}
\begin{tabular}{llllll} 
\hline
\multirow{2}{*}{Type}            & \multirow{2}{*}{Method}                          & \multicolumn{2}{l}{Market1501} & \multicolumn{2}{l}{DukeMTMC}   \\ 
\cline{3-6}
                                 &                                                  & mAP           & Rank-1         & mAP           & Rank-1         \\ 
\hline
\multirow{2}{*}{Pose-guided}     & PGFA \cite{2019Pose}                   & 76.8          & 91.2           & 65.5          & 82.6           \\
                                 & HOReI \cite{2020High}                   & 84.9          & 94.2           & 75.6          & 86.9           \\ 
\hline
\multirow{4}{*}{Part-feature}    & BDB \cite{2018Batch}                   & 84.3          & 94.2           & 72.1            & 86.8             \\
                                 & Pyramid\cite{2018A}                    & 88.2          & 95.7           & 79            & 89             \\
                                 & MGN\cite{2018Learning}               & 86.9          & 95.7           & 78.4          & 88.7           \\
                                 & AANet-152 \cite{2020AANet}             & 83.41         & 93.93          & 74.29        & 87.65           \\
                                 & ISP \cite{2020Identity}                 & 88.6          & 95.3           & 80            & 89.6           \\ 
\hline
\multirow{3}{*}{Attention} & ABD-Net\cite{2019ABD}                  & 88.2          & 95.6           & 78.5          & 89             \\
                                 & RGA-SC \cite{2020Relation}              & 88.1          & 95.8           & 74.9          & 86.1           \\
                                 & Mancs \cite{2018Mancs}                  & 82.3          & 93.1           & 71.8          & 84.9           \\ 
\hline
\multirow{2}{*}{Other}           & Baseline \cite{2019Bags}        & 85.9          & 94.5           & 76.4          & 86.4           \\
                                 & CAGCN \cite{2020Context}                & \textbf{91.7} & \textbf{95.9}  & \textbf{85.9} & \textbf{91.3}  \\ 
\hline
\multirow{2}{*}{\textbf{Ours}}   & \textbf{PGANet-50}                               & 88.5          & 95.4           & 78.8          & 89.4           \\
                                 & \textbf{PGANet-152}                              & 89.3          & 95.4           & 79.6          & 89.1           \\
\hline
\end{tabular}
\end{table}

\subsection{Comparison With State-of-the-Art Methods}

\textbf{PGANet versus Other Model.} To prove the effectiveness of PGANet. We compare it with several state-of-the-art models in person re-identification task. As shown in Table \ref{tab:compare}, our model achieves competitive performance in rank stage. We tested the effects of PGANet using ResNet50 and ResNet152 as backbones respectively. Compared with the attention-based model, PGANet mixes graph structure and attention to optimize features, thus can obtain hidden high-order information and highlight important features well. Compared with part-based models, in PGANet, global features can be obtained to achieve competitive performance. With the help of the part-based method, we can achieve better performance. Without any additional models, PGA module can more easily obtain important features than Pose-guided models. CAGCN can obtain better performance by using probe-gallery relations with the help of picture-wise GCN. So we can draw the conclusion that GCN has great potential in the field of computer vision.

\begin{table}[!htb]
\centering
\caption{Performance (\%) comparisons with the state-of-the-arts methods on Occluded-Duke.}
 \label{occ}
\begin{tabular}{ccc} 
\hline
     \multirow{2}{*}{Method}                        & \multicolumn{2}{l}{Occluded-Duke}\\
                                                    & mAP           & Rank-1\\
\hline
    Ad-Occluded \cite{2018Adversarially} (CVPR2018)   &32.2           &44.5   \\
    PGFA \cite{2019Pose} (ICCV2019)                   &37.3           &51.4   \\
    HOReID \cite{2020High} (CVPR2020)                  &43.8           &55.1   \\
\hline                      
    \textbf{PGANet-50}                              & 50.5          &59.7  \\
    \textbf{PGANet-152}                             & \textbf{54.8}          &\textbf{65.2}  \\

\hline
\end{tabular}
\end{table}
\textbf{Experiment on Occluded-Duke} As shown in Table\ref{occ}, PGANet achieves mAP 54.8\% and Rank-1 65.2\% on Occluded-Duke. Compared with HOReID, it has an increase of 11 points. The results show that PGANet can give higher attention to important features, thus obtain more robust representation even in complex scenes

\begin{table}[htb]
\centering
\caption{Performance (\%) comparisons with the state-of-the-arts in rerank}
\label{rr}
 \setlength{\tabcolsep}{1mm}{
\begin{tabular}{lllll} 
\hline
    \multirow{2}{*}{Method}                          & \multicolumn{2}{l}{Market1501} & \multicolumn{2}{l}{DukeMTMC}   \\ 
                                                     &mAP           & Rank-1         & mAP           & Rank-1         \\ 
\hline
Strong Baseline+RR \cite{2019Bags} (CVPR2019)      &94.2           &95.2            &89.1           &90.3\\
AANet-152+RR \cite{2020AANet} (CVPR2020)            &92.38          &95.1            &86.87          &90.36\\
CAGCN+RR \cite{2020Context} (AAAI2021)                &94.0           &96.2            &88.9           &\textbf{92.3}\\
\hline                      
\textbf{PGANet-50+RR}                               &94.6           &96.0            &89.9           &91.5\\
\textbf{PGANet-152+RR}                              &\textbf{94.8}           &\textbf{96.2}            &\textbf{90.0}           &92.0\\

\hline
\end{tabular}}
\end{table}

\textbf{Re-rank experiment} As shown in Table \ref{rr}, our model also achieves excellent performance (mAP 94.6\% and Rank-1 96.0\%). In the re-rank stage, the k-reciprocal method \cite{2017Re} is adopted as post-processing in our model, which optimizes the results by making effective use of the relationship between the ranking results and the query images. Why does our model achieve state-of-the-art performance after using the re-rank strategy? We compare our model with the best model CAGCN in Table \ref{tab:compare}. The core idea of  CAGCN is to take advantage of the probe-gallery relations, which is consistent with the idea of k-reciprocal method. It essentially uses information between pictures. Our model focuses on optimizing the features of images and ignores the relationship between images. Therefore, Once the relationship between pictures is also used, the model effect has been greatly improved. The experimental results also prove that our model can obtain more robust features.

\subsection{Ablation Study}

\begin{figure}[htb]

  \centering
  \includegraphics[width=\linewidth]{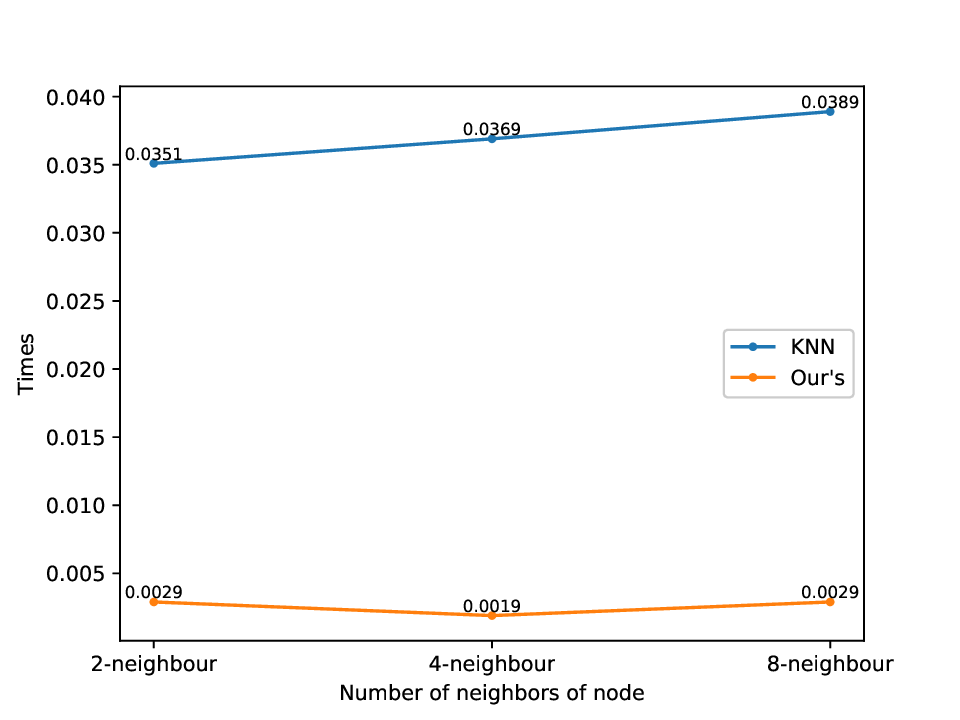}
  \caption{The speed of algorithm.}
  \Description{}
  \label{speed}
\end{figure}

\textbf{Advantages of Graph Generation algorithm} To test the efficiency of our algorithm, we compare the algorithm with the traditional KNN based algorithm. The size of the feature map used for testing is [16,8,512]. We evaluate the time required to create three graphs in which nodes have different numbers of neighbors, ie. 2-neighbor, 4-neighbor, and 8-neighbor. In 2-neighbor, the graph is built on the channel of the picture, and in 4-neighbor and 8-neighbor, is built on the pixels of the picture. The values are obtained as the average of running time of the algorithm, and drawn as a line graph. As shown in Fig. \ref{speed}, the algorithm proposed in this paper increases the speed of graph generation by an order of magnitude, at least 12 times faster than a traditional KNN based algorithm.

\begin{figure}[htb]
  \centering
  \includegraphics[width=\linewidth]{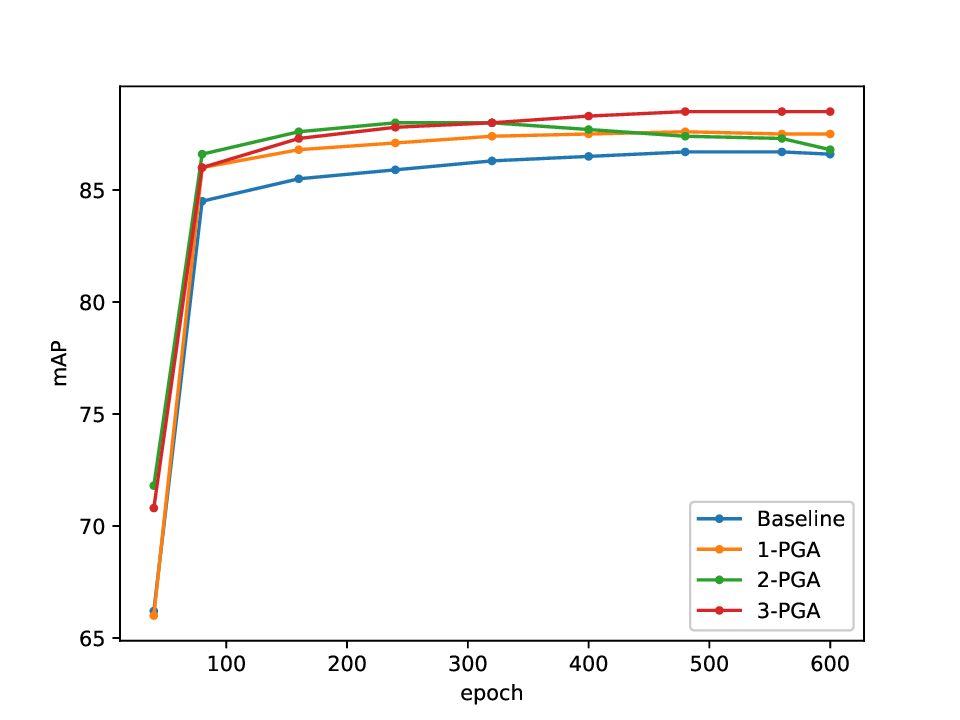}
  \caption{The effect of different layers of multi-layer PGA on the results.}
  \Description{}
  \label{layer}
\end{figure}

\begin{figure}[H]
  \centering
  \includegraphics[width=\linewidth]{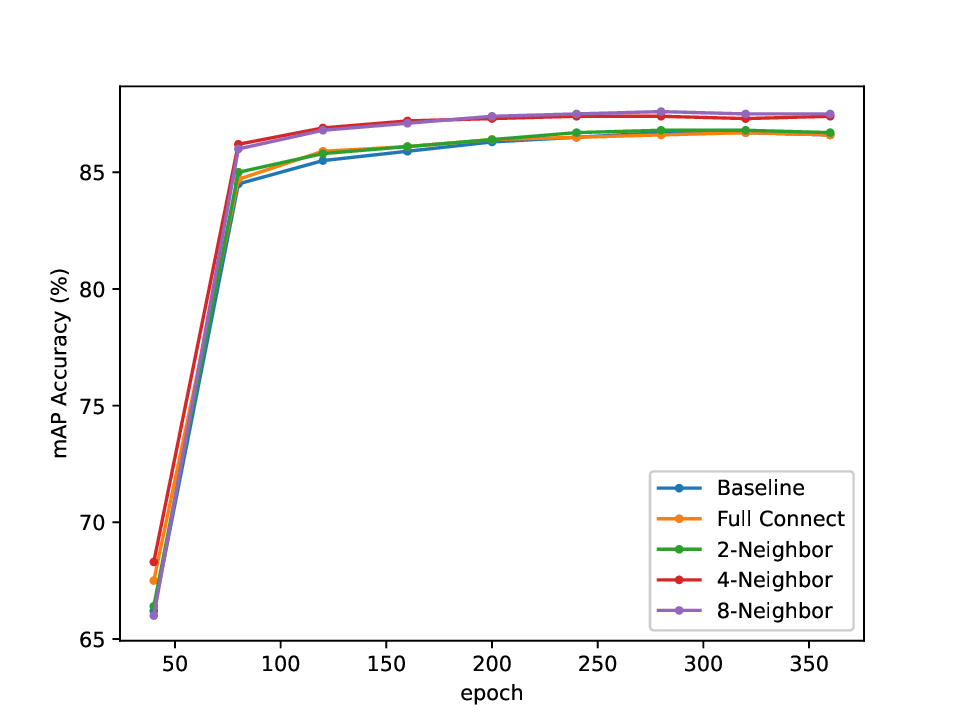}
  \caption{The impact of using different graphs on the model.}
  \Description{}
  \label{neibour}
\end{figure}

\textbf{Advantage of multi-layer PGA} In this section, we test the influence of the number of layers of PGA on the model. Fig. \ref{layer} shows continuous enhancement of the effect with the increase of the number of PGA layers. Compared with strong baseline, the mAP of the 3-layer model has increased by 2.6\%, and the accuracy of Rnk-1 has increased by 0.9\% (based on ResNet50). The experimental results show that the larger receptive field effectively improves the model's representation ability and can obtain hidden high-level information as the number of PGA layers increases. As shown in Fig. \ref{layer} and \ref{neibour}, even if the model has been trained for many rounds, we can not find the phenomenon of vanishing gradient and over smooth. It is noticeable that PGA with residual structure can effectively alleviate the vanishing gradients problem.

\textbf{Influence of neighbors} In PGA, graph convolution to optimize feature maps for better representation. In fully connected graphs, the non local method is used  for nodes to obtain the one-to-all relationship. As shown in Fig. \ref{neibour}, better performance is gotten since structural information can be obtained by the model, with the help of the better sparse and strict local graph produced by our algorithm.

\section{Conclusion}
In this paper, we explored the application of GCN in the image feature extraction stage. A new image feature extraction framework PGANet is proposed, which can obtain a larger receptive field and make good use of structural information to obtain hidden high-order information. PGANet uses an alternate structure of CNN and multi-layer PGA.CNN is used to extract image features, multi-layer PGA is used for the obtained feature maps to optimize features and expand the receptive field. Therefore, PGANet can get a more robust representation of the picture. To apply GCN in the feature extraction stage of the picture, a brand-new graph generation algorithm is proposed which can efficiently generate a sparse and strictly local graph corresponding to the picture. Combine our algorithm with the modified GAT, we get the PGA module. With iteration of\ PGA, high-ord information can be obtained. Extensive experiments on person re-identification demonstrate the effectiveness of our proposed framework.

\begin{acks}
This work was supported by the National Key Research and Development Program of China under Grant 2020YFB2103803.
\end{acks}

\bibliographystyle{ACM-Reference-Format}
\balance
\bibliography{sample-base}

\end{document}